\documentclass[letterpaper, 10 pt, conference]{ieeeconf}  % Comment this line out if you need a4paper

\usepackage[utf8]{inputenc} % allow utf-8 input
\usepackage[T1]{fontenc}    % use 8-bit T1 fonts
\usepackage{hyperref}       % hyperlinks
\usepackage{url}            % simple URL typesetting
\usepackage{booktabs}       % professional-quality tables
\usepackage{amsfonts}       % blackboard math symbols
\usepackage{nicefrac}       % compact symbols for 1/2, etc.
\usepackage{microtype}      % microtypography
\usepackage{xcolor}         % colors
\usepackage{amsmath}
\usepackage{amssymb}
\usepackage{mathtools}
\usepackage{algorithm}
\usepackage{algorithmic}
\usepackage{wrapfig}

\usepackage[capitalise]{cleveref}
\usepackage{amsmath,amsfonts,amssymb}
\usepackage{mathtools}

\newtheorem{proposition}{Proposition}

\newtheorem{assumption}{Assumption}
\usepackage{svg}

\IEEEoverridecommandlockouts                              % This command is only needed if 
\newcommand{\data}{\mathcal{D}}

\newcommand{\datatr}{\mathcal{D}_{tr}}
\newcommand{\datacal}{\mathcal{D}_{cal}}

\newcommand{\quantile}[1]{\text{Quantile}_{#1}}
\newcommand{\alphalow}{\alpha_{\text{lo}}}
\newcommand{\alphahigh}{\alpha_{\text{hi}}}
\newcommand{\technicalreport}{}

\title{\LARGE \bf
Conformal Off-Policy Evaluation in Markov Decision Processes
}

\author{Daniele Foffano$^\ast$, Alessio Russo$^\ast$ and Alexandre Proutiere% <-this % stops a space
\thanks{$\ast$ Equal contribution}% <-this % stops a space
\thanks{Daniele Foffano, Alessio Russo and Alexandre Proutiere are in the Division of Decision and Control Systems of the EECS School at KTH Royal Institute of Technology, Stockholm, Sweden.
        {\tt\small \{foffano,alessior,alepro\}@kth.se}}%
}

\begin{document}

\maketitle
\thispagestyle{empty}
\pagestyle{empty}

\begin{abstract}
    Reinforcement Learning aims at identifying and evaluating efficient control policies from data. In many real-world applications, the learner is not allowed to experiment and cannot gather data in an online manner (this is the case when experimenting is expensive, risky or unethical). For such applications, the reward of a given policy (the {\it target} policy) must be estimated using historical data gathered under a different policy (the {\it behavior} policy). Most methods for this learning task, referred to as Off-Policy Evaluation (OPE), do not come with accuracy and certainty guarantees. We present a novel OPE method based on Conformal Prediction that outputs an interval containing the true reward of the target policy with a prescribed level of certainty. The main challenge in OPE stems from the distribution shift due to the discrepancies between the target and the behavior policies. We propose and empirically evaluate different ways to deal with this shift. Some of these methods yield conformalized intervals with reduced length compared to existing approaches, while maintaining the same certainty level.
\end{abstract}
\section{Introduction}

In this work, we consider the problem of off-policy evaluation (OPE) in finite time-horizon Markov Decision Processes (MDPs). This problem is concerned with the task of learning the expected cumulative reward of a {\it target} policy from data gathered under a different {\it behavior} policy. In fact, OPE has attracted a lot of attention recently \cite{le2019batch, shi2022statistical,precup2000eligibility, thomas2015highevaluation, hanna2017bootstrapping, liu2018breaking} since it is particularly relevant in real-world scenarios where the learner is not allowed to experiment and deploy the target policy to infer its value. In these scenarios, testing a new policy in an online manner can be indeed too risky or unethical (e.g., in finance or healthcare).

The main challenge in OPE algorithms stems from the distribution shift of the target and behavior policies. To address this issue, researchers have developed various solutions, often based on Importance Sampling methods (refer to \textsection \ref{sec:related} and to \cite{uehara2022review} for a recent survey). Lastly, while existing OPE algorithms sometimes enjoy asymptotic convergence properties, most of them do not come with accuracy and certainty guarantees \cite{thomas2015highevaluation,thomas2015highimprovement,kallus2020double}.

To that aim, we are concerned with devising OPE estimators that enjoy non-asymptotic performance guarantees. We leverage techniques from Conformal Prediction (CP) \cite{vovk2005algorithmic,tibshirani2019conformal,romano2019conformalized}, which, directly from the data, allow to build {\it conformalized} sets that provably includes the true value of the quantity to be estimated with a prescribed level of certainty. Furthermore, CP is a distribution-free method, thus circumventing the burden of estimating a model while providing non-asymptotic guarantees. Due to these desirable properties, CP has been applied with success in many fields, including medicine \cite{lindh2017predicting, zhan2020electronic, lu2022three}, aerospace engineering \cite{xi2022conformal}, finance \cite{wisniewski2020application} and safe motion planning \cite{lindemann2022safe}.

Nevertheless, standard CP assumes to be trained on i.i.d. data, and that  at test time the data comes from the same distribution from which the training data was drawn (a.k.a. as {\it distribution/covariates shift}). 
This latter assumption is violated in OPE problems, since the training data is gathered using a policy than is different from  the target policy to be evaluated.
A solution to address the distribution shift  is to leverage the concept of {\it weighted exchangeability} \cite{tibshirani2019conformal, lei2021conformal}.

By exploiting the concept of weighted exchengeability, we study the \emph{conformalized OPE} problem for Markov Decision Processes (MDPs). Our method builds on top of the technique described in \cite{taufiq2022conformal}, which introduces conformalized OPE for contextual bandit models  (which can be seen as MDPs with i.i.d. states). Compared to \cite{taufiq2022conformal},  we have to handle additional difficulties, including the inherent dependence in the data (which consists of trajectories of a controlled Markov chain) and the statistical hardness of dealing with the distribution shift when the time horizon grows large.

{\it Contribution-wise,} we present and empirically evaluate CP algorithms that yield conformalized intervals with reduced length compared to existing approaches, while maintaining the same certainty level. These algorithms are based on the two following new components. (i) Asymmetric score functions: existing CP approaches use symmetric score functions and hence, for our problem, would output conformalized intervals centered on the value of the behavior policy. We introduce asymmetric score functions, so that the CP algorithm yields an interval that efficiently moves its center to follow the distribution shift. In turn, CP with asymmetric score functions results in intervals of  smaller size. (ii) We propose methods to address the distribution shift in MDPs. 
 
We finally illustrate the performance of our algorithms numerically on the classical inventory control problem \cite{puterman2014markov}. The experiments demonstrate that indeed our algorithms achieve smaller interval lengths than existing approaches, while retaining the same certainty guarantees.

\section{Related work}\label{sec:related}

\subsection{Off-Policy Evaluation (OPE)} 

There are mainly three classes of OPE algorithms in the literature: Direct, Importance Sampling and Doubly Robust Methods. Direct Methods (DMs)  learn a model of the system \cite{le2019batch, shi2022statistical} and then evaluate the policy against it. DMs can lead to biased estimators due to a mismatch between the model and the true system. Importance Sampling (IS) is a well-known method \cite{precup2000eligibility, thomas2015highevaluation, hanna2017bootstrapping, liu2018breaking} used to correct the distribution mismatch caused by the discrepancies between the target and the behavior policies by re-weighting the sampled rewards. Still, IS-based algorithms suffer from high variance in long-horizon problems. Doubly Robust (DR) methods combine DMs and IS to obtain more robust estimators \cite{kallus2020double, jiang2020minimax}. \cite{liu2018breaking} introduce Marginalized Importance Sampling, reducing the variance by applying IS directly on the stationary state-visitation distribution. 

The aforementioned approaches only provide an accurate point-wise estimate of the policy value, without quantifying its uncertainty. 
\cite{bottou2013counterfactual} derived  confidence intervals (CIs) using the Central Limit Theorem. In \cite{thomas2015highevaluation, kuzborskij2021confident}, the authors leveraged concentration inequalities to estimate good CIs, which, however, tend to be overly-conservative. For short-horizon problems, \cite{thomas2015highimprovement, hanna2017bootstrapping} approximate CIs for OPE can also be found by means of  bootstrapping.  \cite{shi2021deeply} derives a non-asymptotic CI using concentration bounds on a kernel-based Q-function. 

In \cite{kallus2022efficiently}, the authors derive an asymptotic CI using Double Reinforcement Learning (DRL), also addressing the curse of the horizon. However, the DRL method might not converge in high-dimensional RL tasks, resulting in an asymptotically biased estimator. \cite{duan2020minimax, shi2022statistical} derive non-asymptotic and asymptotic CIs by approximating the value function with linear functions, but their approaches might lead to a biased estimator if the model assumption is incorrect. \cite{kallus2020double} derived a CI that involves solving a linear program, but they assume the observations to be i.i.d., whereas transitions are time-dependent in many RL problems.

\subsection{Conformal Prediction (CP)} 

CP is a frequentist technique to derive CIs with a specified coverage (\emph{i.e.}, confidence) and a finite  number of i.i.d. samples (we refer the reader to \cite{manokhinvalery20226467205} for a comprehensive list of CP-related papers). The advantage of CP with respect to other methods is that the provided coverage guarantees are distribution-free and non-asymptotic. 

CP for off-policy evaluation has been recently applied to the contextual bandit setting \cite{taufiq2022conformal}, which, in contrast to our work, has no dynamics and no time-dependent data. To address the distribution shift, the authors in \cite{taufiq2022conformal} use of the weighted exchangeability property, which was previously introduced in \cite{tibshirani2019conformal}. In \cite{dietterich2022conformal}, the authors apply CP to predict the expected value of MDPs trajectories. They consider an online setting where they do not have to deal with the distribution shift.

\section{Preliminaries}

\subsection{Off-policy evaluation in Markov Decision Processes}

We consider finite-time horizon MDPs \cite{puterman2014markov}. Such an MDP is defined by a tuple $M=\langle {\cal X}, {\cal A}, T, q, p, H \rangle$, where ${\cal X}$ and ${\cal A}$ are the (finite) state and action spaces, respectively. For all $(x,a)\in {\cal X}\times {\cal A}$, $T(\cdot | x, a)$ and $q(\cdot|x,a)$ denote the distributions of the next state and of the instantaneous reward given that the current state is $x$ and that the decision maker selects action $a$ (for simplicity, we assume that the transition probabilities and the reward distributions are stationary; our results can be easily generalized to non-stationary dynamics and rewards). Finally, $p\in \Delta({\cal S})$ denotes the distribution of the initial state, and $H$ the time horizon.

In off-policy evaluation, we gather data using a behavior policy $\pi^b$, and we wish to estimate the value function of different policy $\pi$. Here again for simplicity, we consider stationary policies: both $\pi^b$ and $\pi$ are mappings between the state space and the set $\Delta({\cal A})$ of distributions over actions. The value function of $\pi$ maps the initial state $x$ to the expected reward gathered under $\pi$ when starting in $x$:  $V_H^\pi(x)=\mathbb{E}_\pi[\sum_{t= 1}^H r_t|x_1=x ]$, where $r_t\sim q(\cdot|x_t,a_t)$,  $a_t\sim \pi(\cdot| x_t)$, and $x_{t+1}\sim T(\cdot| x_t,a_t)$ for $t=1,\ldots,H$.

\subsection{Standard Conformal Prediction}

Conformal Prediction (CP) is a method for distribution-free uncertainty quantification of learning methods, see e.g. \cite{vovk2005algorithmic, papadopoulos2002inductive, lei2014distribution}. To illustrate how CP works, we consider classical supervised learning tasks and restrict our attention to split CP where the pre-training and the calibration phases are conducted on different datasets. The learner starts with a pre-trained model $\hat{f}:{\cal X}\to {\cal Y}$ that maps inputs to predicted labels (this model may also consist of upper and lower estimated quantiles if the pre-training procedure corresponds to quantile regression). She also has i.i.d. calibration data $\datacal= \{X_i, Y_i\}_{i=1}^n \overset{\text{i.i.d.}}{\sim}P_{X,Y}$. From $\hat{f}$ and $\datacal$, CP constructs for each possible input $x$ a subset $\hat{C}_n(x)$ of possible labels. More precisely, the method proceeds as follows: (i) first a score function $s:{\cal X}\times {\cal Y}\to \mathbb{R}$ is constructed from the model $\hat{f}$ (e.g., it could be the residuals $|y-\hat{f}(x)|$ if ${\cal Y}\subset \mathbb{R}$); (ii) the scores of the various calibration samples are computed $V_i = s(X_i,Y_i)$, and (iii) the confidence set is built according to $\hat{C}_n(x)=\{ y\in {\cal Y}: s(x,y) \leq \eta \}$, where $\eta = \quantile{1-\alpha}\left({1\over n+1}\left(\sum_{i=1}^n\delta_{V_i}+\delta_{\{\infty\} }\right)\right)$. If $(X_1,Y_1),\ldots,(X_{n+1},Y_{n+1})$ are exchangeable, this construction ensures coverage with certainty level $1-\alpha$:
\begin{equation}
    1-\alpha \leq \mathbb{P}(Y \in \Hat{C}_n(X)) \leq 1-\alpha +\frac{1}{n+1}.
\end{equation}

\section{Conformalized Off-Policy Evaluation}

\begin{figure}
    \centering
    \includegraphics[width=0.9\linewidth]{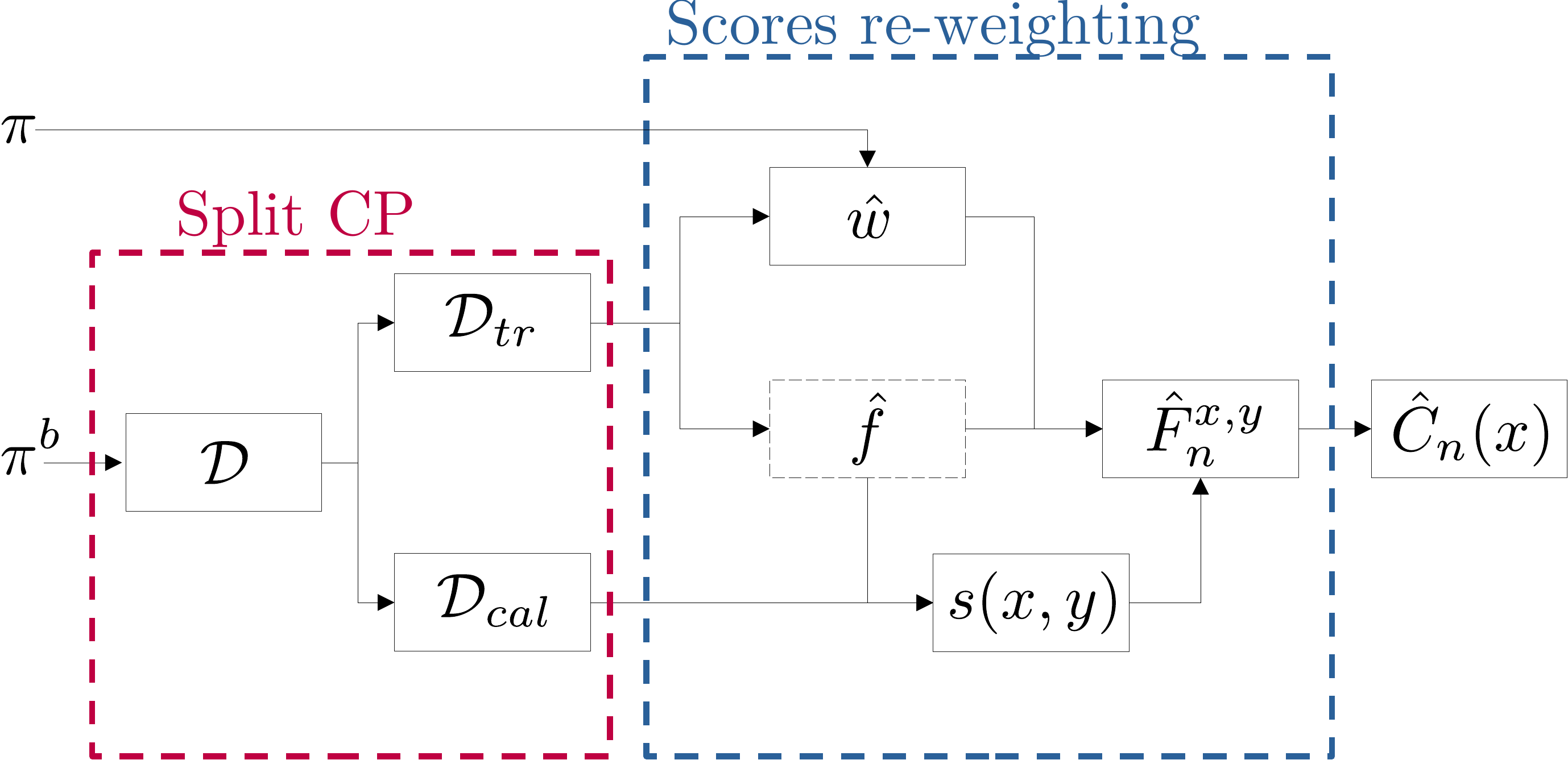}
    \caption{Conformal prediction for off-policy evaluation. The dataset $\data$ is collected using a behavior policy $\pi^b$, which is then split into the \textit{training} $\datatr$ and \textit{calibration} $\datacal$ datasets. When evaluating a different policy $\pi$, there is a shift in the data distribution, and we need to learn a likelihood ratios $\hat w$ to compensate for this shift. The training data is used to learn estimates of the weights $\hat{w}$ and a model $\hat{f}$ used in the computation of the scores. The estimated weights are used as plug-in estimates to re-weight the cumulative distribution function of the scores $\hat{F}_n^{x,y}$, which is then used to compute the conformalized intervals $\hat C_n(x)$.}
    \label{fig:OPE_diagram}
\end{figure}

Our objective is to get conformalized predictions for the value function of a policy $\pi$, based on training and calibration data gathered under a different behavior policy $\pi_b$. We address this {\it distribution shift} by extending and improving the techniques developed in \cite{tibshirani2019conformal,taufiq2022conformal}. We apply the CP formalism where the input $X$ corresponds to the initial state, and the output $Y$ to $V_H^\pi(X)$. Our method is illustrated in Figure \ref{fig:OPE_diagram}. Next, we describe its components in detail. Specifically, (i) we explain how the aforementioned distribution shift can be addressed by weighing scores; (ii) we then discuss the important choice of the score function.  

\subsection{Weighted conformal prediction}

As suggested \cite{tibshirani2019conformal,taufiq2022conformal}, we can handle the distribution shift by weighing the scores using estimates of the likelihood ratio
\[
w(x,y) \coloneqq \frac{{\rm d}P_{X,Y}^{\pi}}{{\rm d}P_{X,Y}^{\pi^b}}(x,y) = \frac{{\rm d}P_{Y|X}^{\pi}}{{\rm d}P_{Y|X}^{\pi^b}}(y|x),
\]
where for any policy $\pi'\in \{\pi,\pi^b \}$, $P_{X,Y}^{\pi'}(x,y)=P_{Y|X}^{\pi'}(y|x)p(x)$ denotes the distribution of the observation $(X,Y)$ under $\pi'$ ($P_{Y|X}^{\pi'}$ is this distribution given $X$), and $p(x)$ is the initial state distribution, which is the same in both cases. The value of a given trajectory $\tau=\{x_1,a_1,r_1,\dots,x_H,a_H,r_H \}$ is $y=\sum_{t=1}^Hr_t$. For any policy $\pi'\in \{\pi,\pi_b \}$, the probability of observing $\tau$ under $\pi'$ given the initial state $x_1=x$ is:
\begin{align*}
P^{\pi'}(\tau|x)= & \pi'(a_1|x)q(r_1|x,a_1)\prod_{t=2}^H \pi'(a_t|x_t)\\
&\qquad\times T(x_{t}|x_{t-1},a_{t-1})q(r_{t}|x_{t},a_{t}).
\end{align*}
Hence the weights can be written as:
\[
w(x,y) = \frac{\int \mathbf{1}_{\{ y= \sum_{t=1}^Hr_t\} } P^\pi(\tau|x) {\rm d}\tau}{\int \mathbf{1}_{\{ y= \sum_{t=1}^Hr_t\} } P^{\pi^b}(\tau|x) {\rm d}\tau}.
\]
We make the following assumption to guarantee that the above weights are always well defined, and that the calibration data is i.i.d. 
\begin{assumption}\label{assumption:assumption1}{\it
We assume throughout the paper that $P^{\pi}(\cdot|x)$ is absolutely continuous w.r.t. $P^{\pi^b}(\cdot | x)$ for all $x\in {\cal X}$. We further assume that calibration data $\datacal$ provides $n$ i.i.d. samples $(X_i,Y_i)\sim P_{X,Y}^{\pi^b}$. }
\end{assumption}
Then, we can compute the scores $V_i=s(X_i,Y_i)$. For each possible pair $(x,y)$, using the normalized weights, we form the distribution $
\hat{F}_n^{x, y}\coloneqq \sum_{i=1}^{n} p_{i}^{w}(x, y) \delta_{V_{i}}+p_{n+1}^{w}(x, y) \delta_{\infty}$, with 
\begin{equation}\label{eq:weighted_probabilities_cdf}
p_{i}^{w}(x, y)= \begin{cases}
\dfrac{w\left(X_{i}, Y_{i}\right)}{\sum_{j=1}^{n} w\left(X_{j}, Y_{j}\right)+w(x, y)}\quad \hbox{if }i\leq n,\\
\dfrac{w\left(x,y\right)}{\sum_{j=1}^{n} w\left(X_{j}, Y_{j}\right)+w(x, y)}\quad \hbox{if }i=n+1,
\end{cases}
\end{equation}
and the conformalized set 
\begin{equation}\label{eq:confidence_set_single_quantile}
    \hat{C}_n(x)\coloneqq\left\{y \in \mathbb{R}: s(x, y) \leq \quantile{1-\alpha}\left(\hat{F}_n^{x, y}\right)\right\}.
\end{equation}

\begin{proposition}\label{prop:original_coverage}
{\it
Under  \cref{assumption:assumption1}, for any score function $s$ and any $\alpha \in (0,1)$, 
\begin{equation}
    \mathbb{P}^{\pi^b,\pi}\big[ Y \in \hat{C}_n(X)\big] \geq 1-\alpha,
\end{equation}
where $\mathbb{P}^{\pi^b,\pi}$ accounts for the randomness of $(X, Y)\sim P_{X, Y}^{\pi}$ and that of the data $\datacal = \{X_i,Y_i\}_{i=1}^n$ (with for all $i\in [n]$, $(X_i,Y_i)\sim P_{X,Y}^{\pi^b}$). 
}
\end{proposition}

\medskip
%#######################################
\ifdefined \technicalreport
\begin{proof}
The proof follows that in \cite[Proposition 4.1]{taufiq2022conformal}. The idea relies on the fact that $\{(X_i,Y_i)\}_{i=1}^{n}\cup (X_{n+1}, Y_{n+1})$ are weighted exchangeable (see Lemma 1 in the appendix), where $(X_{n+1}, Y_{n+1})$ is sampled according to $P_{X,Y}^\pi$ and $\{(X_i,Y_i)\}_{i=1}^{n}$ according to $P_{X,Y}^{\pi^b}$.
Then, assume for simplicity that $V_{1}, \ldots, V_{n+1}$ are distinct almost surely. We define $f$ as the joint distribution of the random variables $\left\{X_{i}, Y_{i}\right\}_{i=1}^{n+1}$. We also denote $E_{z}$ as the event of $\left\{Z_{1}, \ldots, Z_{n+1}\right\}=\left\{z_{1}, \ldots, z_{n+1}\right\}$ (where the equality refers to the equality between sets) and let $v_{i}=s\left(z_{i}\right)=s\left(x_{i}, y_{i}\right)$. Then, for each $i$ :
\begin{align*}
    &\mathbb{P}\left[V_{n+1} =v_{i} \mid E_{z}\right]=\mathbb{P}\left[Z_{n+1}=z_{i} \mid E_{z}\right],\\
    &=\frac{\sum_{\sigma: \sigma(n+1)=i} f\left(z_{\sigma(1)}, \ldots, z_{\sigma(n+1)}\right)}{\sum_{\sigma} f\left(z_{\sigma(1)}, \ldots, z_{\sigma(n+1)}\right)}
\end{align*}
Now using the fact that $Z_{1}, \ldots, Z_{n+1}$ are weighted exchangeable, as in \cite{taufiq2022conformal} we find that $\mathbb{P}\left[Z_{n+1}=z_{i} \mid E_{z}\right]=p_{i}^{w}\left(z_{n+1}\right).$

Next, just as in \cite{tibshirani2019conformal} we can view:
\[
\{V_{n+1}=v_{i} \mid E_{z}\} \sim \sum_{i=1}^{n+1} p_{i}^{w}\left(z_{n+1}\right) \delta_{v_{i}}
\]
which implies that:
\[
\mathbb{P}\left[V_{n+1} \leq \quantile{1-\alpha}\left(\sum_{i=1}^{n+1} p_{i}^{w}\left(X_{n+1}\right) \delta_{v_{i}}\right) \mid E_{z}\right] \geq 1-\alpha .
\]
Marginalizing over $E_z$ concludes the proof.
\end{proof}
\else
The proof of the proposition is similar to that of \cite[Proposition 4.1]{taufiq2022conformal}, and is omitted due to space constraints. The complete proofs of all results can be found in the companion technical report\footnote{Find the technical report and the code here \url{https://github.com/danielefoffano/Conformal_OPE_MDP/blob/main/Conformal_OPE_in_MDP.pdf}}. 
\fi
Proposition \ref{prop:original_coverage} shows that, in absence of data from the target policy, we can still use a shifted CDF of the scores to assess the target policy. The result however relies on the assumption that the weights $w(x,y)$ are known. In practice, we could use the training data to learn these weights, refer to Section \ref{sec:offline_est} for details. The next proposition quantifies the impact of the error in this estimation procedure on the coverage. Its proof follows the same arguments as those in \cite{taufiq2022conformal}.

\begin{proposition}\label{proposition:uncertain_weight_single_quantile}{\it 
Assume that the conformalized sets (\ref{eq:confidence_set_single_quantile}) are defined using estimated the weights $\hat w(x,y)$ satisfying $\mathbb{E}^{\pi^b}\left[\hat{w}(X, Y)^r\right]\leq M_r^r<\infty$ for some $r \geq 2$. Define $\Delta_w = \frac{1}{2} \mathbb{E}^{\pi^b}|\hat{w}(X, Y)-w(X, Y)| $. Then
\begin{equation}
    \mathbb{P}^{\pi^b,\pi}\left[Y \in \hat{C}_n(X)\right]\geq 1-\alpha-\Delta_w,
\end{equation} 
If, in addition, the non-conformity scores $\left\{V_i\right\}_{i=1}^n$ have no ties almost surely, then we also have
\[
\mathbb{P}^{\pi^b,\pi}\left[Y \in \hat{C}_n(X)\right] \leq 1-\alpha+\Delta_w+c n^{1 / r-1},
\]
for some positive constant $c$ depending on $M_r$ and $r$  only.}
\end{proposition} 
%######################################################
\ifdefined \technicalreport
\begin{proof}
    The proof is omitted for brevity, since it follows \emph{mutatis mutandis} from that in \cite[Proposition 4.2]{taufiq2022conformal}.
\end{proof}
\else

\fi
%######################################################

\subsection{Selecting the score function}\label{subsec:score_function}
\label{subsec:score_function}

The choice of the score function critically impacts the size and center of the conformalized sets $\hat{C}_n(x)$. In previous work \cite{romano2019conformalized,taufiq2022conformal}, the pre-training procedure outputs some estimated quantiles $q_{\alphalow}(x)$ and $q_{\alphahigh}(x)$ for the value of the behavior policy with initial state $x$, and the use of the symmetric score function 
\begin{equation}\label{eq:orginal_score}
s(x,y) = \max(q_{\alphalow}(x) - y, y - q_{\alphahigh}(x)),
\end{equation}
is advocated. This choice yields a set $\hat{C}_n(x)$ centered $\bar{q}_{\pi^b}(x)=(q_{\alphalow}(x) + q_{\alphahigh}(x))/2$. Indeed, in view of (\ref{eq:confidence_set_single_quantile}) and (\ref{eq:orginal_score}), there is $\eta(x)\in \mathbb{R}$ such that $\hat{C}_n(x)=[\bar{q}_{\pi^b}(x)-\eta(x), \bar{q}_{\pi^b}(x)+\eta(x)]$ (note that when $n$ grows large, $\eta(x)$ becomes independent of $x$). Having $\hat{C}_n(x)$ centered on the estimated median value for $\pi^b$ is of course very problematic when the values of $\pi^b$ and $\pi$ significantly differ. In this case, the length of $\hat{C}_n(x)$ becomes unnecessarily large. Next we propose methods and score functions that efficiently re-center $\hat{C}_n(x)$ around the value of $\pi$ (instead of $\pi_b$), and that in turn yield much smaller conformalized sets.

\subsubsection{Double-quantile score}\label{paragraph:double_quantile} a first idea is to break the symmetry of the score function used in \cite{taufiq2022conformal} by considering the following confidence set
\begin{align}
    \hat{C}_n(x)&\coloneqq\left\{y \in \mathbb{R}: q_{\alphalow}(x) -y \leq \quantile{1-\alpha/2}\left(\hat{F}_{n,0}^{x, y}\right)\right\}\nonumber\\
    &\quad \cap \left\{y \in \mathbb{R}: y-q_{\alphahigh}(x) \leq \quantile{1-\alpha/2}\left(\hat{F}_{n,1}^{x, y}\right)\right\} \label{eq:confidence_set_double_quantile},
\end{align}
where
$ \hat{F}_{n,0}^{x, y}\coloneqq \sum_{i=1}^{n} p_{i}^{w}(x, y) \delta_{V_{i,0}}+p_{n+1}^{w}(x, y) \delta_{\infty}$ and $\hat{F}_{n,1}^{x, y}\coloneqq \sum_{i=1}^{n} p_{i}^{w}(x, y) \delta_{V_{i,1}}+p_{n+1}^{w}(x, y) \delta_{\infty}$, 
with $V_{i,0} = q_{\alphalow}(X_i) -Y_i$ and  $V_{i,1} =  Y_i-q_{\alphahigh}(X_i) $.
\begin{figure}[t]
    \centering
    \includegraphics[width=0.8\linewidth]{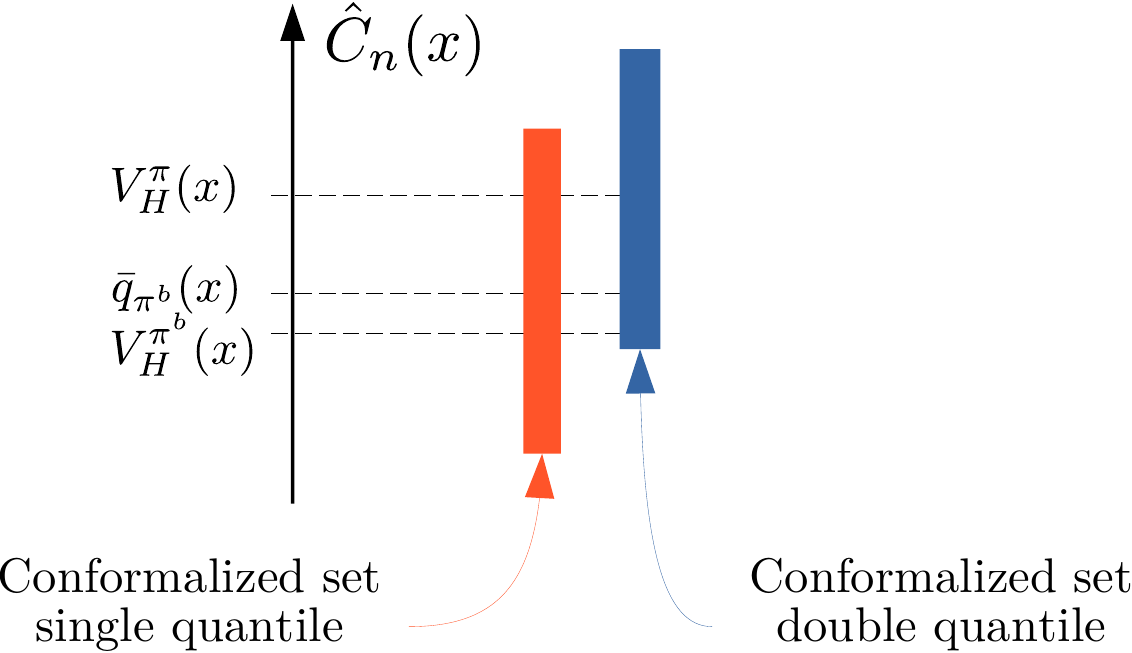}
    \caption{Symmetry problem. For the original confidence set with one single quantile, and score function $s(x,y)=\max(q_{\alphalow}(x) - y, y - q_{\alphahigh}(x))$, we obtain a set that is symmetric around its middle point $(q_{\alphalow}(x) + q_{\alphahigh}(x))/2$. We can break this symmetry by considering two different score quantiles, one for $q_{\alphalow}(x) - y$ and one for $y - q_{\alphahigh}(x)$, thus leading to a less conservative conformalized set.}
    \label{fig:double_quantile_score}
\end{figure}
In essence, we separately look at the lower and upper quantiles of the shifted distribution of the scores. A  graphical illustration is provided in \cref{fig:double_quantile_score}. The new construction of $\hat{C}_n(x)$ does not affect coverage guarantees: 

\begin{proposition}\label{prop:coverage_true_w_double_quantile_score}{\it
    Under \cref{assumption:assumption1}, for $\alpha\in(0,1)$ the sets $\hat C_n(x)$ in (\ref{eq:confidence_set_double_quantile}) satisfies 
    \begin{equation}
        \mathbb{P}^{\pi^b,\pi}\left[Y \in \hat{C}_n(X)\right] \geq 1-\alpha.
    \end{equation}
    }
\end{proposition}
%######################################################
\ifdefined \technicalreport
\begin{proof}
The proof follows from that of Proposition 1. Assume for simplicity that for a fixed $j\in\{0,1\}$ the values $\{V_{i,j}\}_{i=1}^n$ are distinct almost surely and let $s_0(x,y)=q_{\alphalow}(x) -y, s_1(x,y)= y-q_{\alphahigh}(x)$. As before, we define $f$ as the joint distribution of the random variables $\left\{X_{i}, Y_{i}\right\}_{i=1}^{n+1}$. Recall that $Z_i=(X_i,Y_i)$, then, we denote by  $E_{z}$ the event that $\left\{Z_{1}, \ldots, Z_{n+1}\right\}$ $=\left\{z_{1}, \ldots, z_{n+1}\right\}$ and let $v_{i,j}=s_j\left(z_{i}\right)=s_j\left(x_{i}, y_{i}\right)$ for $j\in\{0,1\}$.
First, following the previous proposition, we observe that
\begin{equation}
\{V_{n+1,j}=v_{i,j} \mid E_{z}\} \sim \sum_{i=1}^{n+1} p_{i}^{w}\left(z_{n+1}\right) \delta_{v_{i,j}},\quad j\in \{0,1\},
\end{equation}
and, similarly,
$
\mathbb{P}\left[V_{n+1,j} \leq \eta_{1-\alpha/2}(Z_{n+1}) \mid E_{z}\right] \geq 1-\alpha/2,\quad j\in\{0,1\},
$
where 
\[
\eta_{1-\alpha/2}(Z_{n+1}) = \quantile{1-\alpha/2}\left(\sum_{i=1}^{n+1} p_{i}^{w}\left(Z_{n+1}\right) \delta_{v_{i,j}}\right).
\]
Let  $\hat C_{n,j}=\left\{y \in \mathbb{R}: s_{j}(x,y)\leq  \quantile{1-\alpha/2}\left(\hat{F}_{\eta,j}^{x, y}\right)\right\}$, then
$
 \mathbb{P}^{\pi^b,\pi}\left[Y \notin \hat{C}_{n,j}(X) |E_z\right] \leq \alpha/2, j\in \{0,1\},
$
from which follows (through a union bound)
$
 \mathbb{P}^{\pi^b,\pi}\left[Y \notin \hat{C}_{n}(X) |E_z\right] \leq \alpha.
$
We get the claim after marginalizing over $E_z$.
\end{proof}
\else

\fi
%######################################################
We also obtain the following guarantees in case $w(x,y)$ is replaced by $\hat w(x,y)$.
\begin{proposition}\label{proposition:uncertain_w_double_quantile}
{\it 
    Let $\hat{C}_n(x)$ be as in (\ref{eq:confidence_set_double_quantile}) with weights $w(x, y)$ replaced by $\hat{w}(x, y)$. Under the same assumptions as in \cref{proposition:uncertain_weight_single_quantile}, we have
    \[ \mathbb{P}^{\pi^b,\pi}\left[Y \in \hat{C}_n(X)\right]\geq 1-\alpha-\Delta_w .
   \]
    If, in addition, non-conformity scores $\left\{V_{i,0}\right\}_{i=1}^n$ and $\left\{V_{i,1}\right\}_{i=1}^n$ have no ties almost surely, then we also have
    \[
    \mathbb{P}^{\pi^b,\pi}\left[Y \in \hat{C}_n(X)\right] \leq 1-\alpha+\Delta_w+c n^{1 / r-1},
    \]
    for some positive constant $c$ depending only on $M_r$ and $r$.
}
\end{proposition}
%######################################################
\ifdefined\technicalreport
\begin{proof}
We take inspiration from  \cite[Proposition 4.2]{taufiq2022conformal}. For the sake of notation, we denote the test point by $(X',Y')$ instead of $(X_{n+1},Y_{n+1})$.  We also denote by $\hat C_n(x)$ the confidence set 
$
    \hat{C}_n(x)\coloneqq C_{0,n}(x) \cap C_{1,n}(x),
$
with
\begin{align*}
     \hat{C}_{0,n}(x)&\coloneqq\left\{y \in \mathbb{R}: q_{\alphalow}(x) -y \leq \quantile{1-\alpha/2}\left(\hat{F}_{\eta,0}^{x, y}\right)\right\},\\
    \hat{C}_{1,n}(x)&\coloneqq  \left\{y \in \mathbb{R}: y-q_{\alphahigh}(x) \leq \quantile{1-\alpha/2}\left(\hat{F}_{\eta,1}^{x, y}\right)\right\}
\end{align*}
where
$ \hat{F}_{n,0}^{x, y}\coloneqq \sum_{i=1}^{n} p_{i}^{\hat w}(x, y) \delta_{V_{i,0}}+p_{n+1}^{\hat w}(x, y) \delta_{\infty}$ and $\hat{F}_{n,1}^{x, y}\coloneqq \sum_{i=1}^{n} p_{i}^{w}(x, y) \delta_{V_{i,1}}+p_{n+1}^{w}(x, y) \delta_{\infty}$, 
with $V_{i,0} $ and  $V_{i,1}$  are as before.

\noindent \underline{We first prove the lower bound}. Let $\tilde P_{X,Y}^\pi$ be a probability measure with ${\rm d} \tilde P_{X,Y}^\pi(x,y)=\hat w(x,y) {\rm d}P_{X,Y}^{\pi^b}(x,y)$. Further, let ${\rm TV}(P,Q)$ be the total variation distance between two distributions $P,Q$, and observe that
\[
{\rm TV}(\tilde P^{\pi}, P^{\pi}) = \frac{1}{2}\int |\hat w(x,y)  - w(x,y)| {\rm d}P^{\pi^b}(x,y) = \Delta_w.
\]
First, note that from an application of \cref{prop:coverage_true_w_double_quantile_score} we have that  $\mathbb{P}_{(X,Y)\sim \tilde P_{X,Y}^\pi}[(Y\in \hat C_n(X)]\geq 1-\alpha$. 
Then, we can use the total variation to bound the difference in probability
\[
\resizebox{\hsize}{!}{%
        $
        |\mathbb{P}_{(X,Y)\sim \tilde P_{X,Y}^\pi}[Y\in \hat C_n(X)] - \mathbb{P}_{(X,Y)\sim  P_{X,Y}^\pi}[Y\in \hat C_n(X)] | \leq \Delta_w.
        $%      
}
\]
Using the triangle inequality we find the lower bound:
\begin{align*}
\mathbb{P}_{(X,Y)\sim  P_{X,Y}^\pi}[Y\in \hat C_n(X)]\geq 1-\alpha-\Delta_w.
\end{align*}
\noindent \underline{We now prove the upper bound}. The moment assumption on $\hat w(x,y)$ guarantees that $\hat w$ is bounded a.s. under $P^{\pi^b}$. Then, W.l.o.g.,  assume $\mathbb{E}_{(X,Y)\sim P_{X,Y}^{\pi^b}}[\hat w(X,Y)]=1$. As shown  in \cite[Proposition 4.2]{taufiq2022conformal}, we have $\mathbb{E}_{(X,Y)\sim \tilde P_{X,Y}^{\pi}}[\hat w(X,Y)]\leq M_r^2$. Then
\begin{align*}
    \mathbb{P}[Y' \in& \hat C_n(X')] = \mathbb{P}[Y' \in \hat C_{0,n}(X') \cap Y' \in \hat C_{1,n}(X')],\\
    &\leq \min\left[\mathbb{P}(Y' \in \hat C_{0,n}(X') ), \mathbb{P}(Y' \in \hat C_{1,n}(X')) \right].
\end{align*}
The rest of the proof follows  as in  \cite[Proposition 4.2]{taufiq2022conformal}, where we note that
$
\mathbb{P}[Y' \in \hat C_{i,n}(X') ]\leq 1-\alpha + cn^{1/r -1}, i\in\{1,2\},
$
and thus $\mathbb{P}_{(X,Y)\sim\tilde P_{X,Y}^\pi}[Y' \in \hat C_n(X')] \leq 1-\alpha + cn^{1/r -1}$. Using the triangle inequality on
\[
\resizebox{\hsize}{!}{%
        $
        |\mathbb{P}_{(X,Y)\sim \tilde P_{X,Y}^\pi}[Y\in \hat C_n(X)] - \mathbb{P}_{(X,Y)\sim  P_{X,Y}^\pi}[Y\in \hat C_n(X)] | \leq \Delta_w.
        $%      
}
\]
we conclude that $\mathbb{P}_{(X,Y)\sim P_{X,Y}^\pi}[Y' \in \hat C_n(X')] \leq 1-\alpha + \Delta_w +cn^{1/r -1}$. 
\end{proof}
\else

\fi
%######################################################

\medskip
\subsubsection{Shifted values}\label{paragraph:shifted_values} a second idea is to simply shift the values  of the behavior policy $\pi^b$ using the likelihood ratios $w(x,y)$, as one would in important sampling methods. This can be done by simply using $s(x,y)=y$. This choice of score function makes sense intuitively: if we are interested in the value of the target policy $\pi$, then we may look at the shifted distribution of the values of the behavior policy.

We may also combine this choice with the double-quantile idea
and construct $\hat{C}_n(x)$ as
\begin{equation}\label{eq:confidence_set_shifted_values2}
    \hat{C}_n(x)= \hat C_{n,0}(x) \cap \hat C_{n,1}(x),
\end{equation}
where $\hat C_{n,0}= \left\{y \in \mathbb{R}:y \geq \quantile{\alpha/2}\left(\hat{F}_{n}^{x, y}\right)\right\}$ and $\hat C_{n,1}= \left\{y \in \mathbb{R}:y \leq \quantile{1-\alpha/2}\left(\hat{F}_{n}^{x, y}\right)\right\}$.  \cref{prop:coverage_true_w_double_quantile_score,proposition:uncertain_w_double_quantile} also hold for this choice.

\section{Offline estimation of the likelihood ratios}
\label{sec:offline_est}
In this section, we present various ways to estimate the likelihood ratios $w(x,y)$, and discuss their pros and cons.

\subsection{Monte-Carlo method}
To estimate $w(x,y)$, we need to compute $P_{X,Y}^{\pi}(x,y)$ and $P_{X,Y}^{\pi^b}(x,y)$. Recall that the likelihood ratio is equal to
\[
 w(x,y) = \frac{\int\mathbf{1}_{\{ y= \sum_{t=1}^Hr_t\} } P^\pi(\tau|x) {\rm d}\tau}{\int \mathbf{1}_{\{ y= \sum_{t=1}^Hr_t\} } P^{\pi^b}(\tau|x) {\rm d}\tau},
\]
where $\tau$ is a trajectory of length $H$. Since $P^\pi(\tau|x)$ (sim. $P^{\pi^b}(\tau|x)$)  depends on the transition kernel $T$ and the reward distribution $q$, one needs to estimate these distributions from the data. We may proceed as follows:
\begin{enumerate}
    \item We use the training data $\datatr$ to compute an estimate $(\hat T, \hat q)$  of $(T,q)$ (through maximum likelihood).
    \item Compute an estimate of $\hat w(x,y)$ through Monte-Carlo sampling:
    \begin{equation}
    \hat w(x,y) = \dfrac{(1/h)\sum_{k=1}^h  \mathbf{1}_{\{ y= \sum_{t=1}^H r_{t}^{(k)}\} } }{(1/h)\sum_{k=1}^h   \mathbf{1}_{\{ y= \sum_{t=1}^H r_{t}^{(k)'}\} } },
    \end{equation}
    where $r_{t}^{(k)}$ and $r_{t}^{(k)'}$ are sequences of rewards generated, respectively, by starting in $x$ and following $\pi$ and $\pi^b$, and $h$ is the number of Monte Carlo samples. These trajectories are generated using $\hat{T}$ and $\hat{q}$, estimated in the previous step.
\end{enumerate}
This approach has various shortcomings. First it requires us to estimate the model $(T,q)$. Then it forces us to generate a large number of trajectories, which is heavy computationally. Finally, the term $\mathbf{1}_{\{ y= \sum_{t=1}^Hr_t\}}$  is going to be $0$ most of the times. A possible way to alleviate this issue consists in not including the last reward in the trajectory $\tau$. This implies that we replace $\mathbf{1}_{\{ y= \sum_{t=1}^Hr_t\}}$ by $\hat q(y-\sum_{n=1}^{H-1} r_n|x_H, a_H)$. As it turns out, this naive Monte-Carlo method, used with success in simple scenarios (contextual bandits \cite{taufiq2022conformal}), does not work in MDPs.

\subsection{Empirical and gradient-based methods}
Next we present an alternative and more scalable way to estimate the weights $w(x,y)$ from the training dataset $\datatr$. We make use of the following simple rewriting of the likelihood ratio (also suggested in \cite{taufiq2022conformal}):
\begin{align*}
    w(x,y) &= \frac{P_{X,Y}^\pi(x,y) }{P_{X,Y}^{\pi^b}(x,y) },\\
    &= \int \frac{P_{X,Y}^\pi(x,y) }{P_{X,Y}^{\pi^b}(x,y) } \frac{P_{\tau|X,Y}^{\pi^b}(\tau|x,y)}{P_{\tau|X,Y}^{\pi^b}(\tau|x,y)} P_{\tau|X,Y}^{\pi}(\tau|x,y){\rm d}\tau,\\
    &= \int \frac{P_{X,Y,\tau}^\pi(x,y,\tau) }{P_{X,Y,\tau}^{\pi^b}(x,y,\tau) } P_{\tau|X,Y}^{\pi^b}(\tau|x,y) {\rm d}\tau,\\
    &= \mathbb{E}_{\tau \sim P_{\tau|X=x,Y=y}^{\pi^b}}\left[\frac{P_{X,Y,\tau}^\pi(x,y,\tau) }{P_{X,Y,\tau}^{\pi^b}(x,y,\tau) } \right].
\end{align*}
Next, observe that:
\begin{align*}
    \frac{P_{X,Y,\tau}^\pi(x,y,\tau) }{P_{X,Y,\tau}^{\pi^b}(x,y,\tau)} &= \frac{P(y|x,\tau)P^\pi(\tau|x)}{P(y|x,\tau)P^{\pi^b}(\tau|x)}=\frac{  \prod_{t=1}^H \pi(a_t|x_t) }{\prod_{t=1}^H \pi^b(a_t|x_t) }.
\end{align*}
Hence, learning $w$ amounts to learning the following expectation:
\begin{equation}\label{eq:ratio_policies_w}
    w(x,y) = \mathbb{E}_{\tau \sim P_{\tau|X=x,Y=y}^{\pi^b}}\left[\frac{\prod_{t=1}^H \pi(a_t|x_t) }{\prod_{t=1}^H \pi^b(a_t|x_t) } \right].
\end{equation}
To this aim, we propose the following two approaches.

\medskip
\subsubsection{Empirical estimator}\label{paragraph:empirical_estimator} this method applies to the case $x$ and $y$ belong to some finite spaces ${\cal X}$ and ${\cal Y}$ only. In this case, we can directly estimate $w(x,y)$ from the training data $\datatr$ by simply computing
    \begin{equation}
         \hat w(x,y) = \frac{1}{N(x,y)} \sum_{\tau^i\in\datatr (x,y)} \frac{\prod_{t=1}^H \pi(a_t^{(i)}|x_t^{(i)}) }{\prod_{t=1}^H \pi^b(a_t^{(i)}|x_t^{(i)}) },
    \end{equation}
where the training data $\datatr$ consists of $m$ trajectories generated under $\pi_b$, the $i$-th trajectory in this dataset is $\tau_i=(x_t^{(i)},a_t^{(i)},r_t^{(i)})_{t=1}^H$, $\datatr (x,y)$ are trajectories with initial state and the accumulated reward $x$ and $y$, respectively, and $N(x,y)= | \datatr (x,y)|$. When the likelihood ratios are bounded, we can quantify the accuracy of the above estimates using standard concentration results:

\medskip
    \begin{proposition}\label{prop:guarantees_empirical_estimator}{\it
    Let $(\varepsilon,\delta)\in (0,1)$.  Assume  the ratio $\prod_{t=1}^H \pi(a_t|x_t)/\prod_{t=1}^H \pi^b(a_t|x_t)$ to be bounded in $[m,M]$ for all possible trajectories of horizon $H$ generated under $\pi^b$. If $\min_{x,y}N(x,y)\geq \frac{(M-m)^2}{2\varepsilon^2}\ln \frac{2|{\cal X}||{\cal Y}|}{\delta}$, then  
    \[
        \mathbb{P}^{\pi^b}[|\hat w(X,Y)- w(X,Y)|> \varepsilon] <\delta.
    \]
    Furthermore, we also have $\Delta_w\leq \frac{(M-m)|{\cal X}||{\cal Y}|\sqrt{\pi} }{2\sqrt{2\min_{x,y}N(x,y)}}$.
    }
    \end{proposition}
    %######################################################
    \ifdefined \technicalreport
    \begin{proof}
        The proof is a simple application of Hoeffding's inequality:
        $\mathbb{P}[|\hat w(X,Y)- w(X,Y)|> \varepsilon] < \sum_{x,y} 2e^{\frac{-2N(x,y)\varepsilon^2}{ (M-m)^2}}\leq 2|{\cal X}||{\cal Y}| e^{\frac{-2 \varepsilon^2\min_{x,y}N(x,y)}{ (M-m)^2}},$
    where we made use also of a union bound over ${\cal X}\times {\cal Y}$. Then,  if we choose $\min_{x,y}N(x,y)\geq \frac{(M-m)^2}{2\varepsilon^2}\ln \frac{2|{\cal X}||{\cal Y}|}{\delta}$ then $\mathbb{P}[|\hat w(X,Y)- w(X,Y)|> \varepsilon] <\delta.$
    Regarding the inequality, we find that $\mathbb{E}[|\hat w(X,Y)- w(X,Y)|] \leq 2|{\cal X}||{\cal Y}| \int_0^\infty e^{\frac{-2 \varepsilon^2\min_{x,y}N(x,y)}{ (M-m)^2}}{\rm d}\varepsilon =  \frac{(M-m)|{\cal X}||{\cal Y}|\sqrt{\pi} }{\sqrt{2\min_{x,y}N(x,y)}}$, and thus $\Delta_w \leq \frac{(M-m)|{\cal X}||{\cal Y}|\sqrt{\pi} }{2\sqrt{2\min_{x,y}N(x,y)}}$.
    \end{proof}
    \fi
    %######################################################

    \medskip
    \begin{figure}[b]
        \centering
        \includegraphics[width=.9\linewidth]{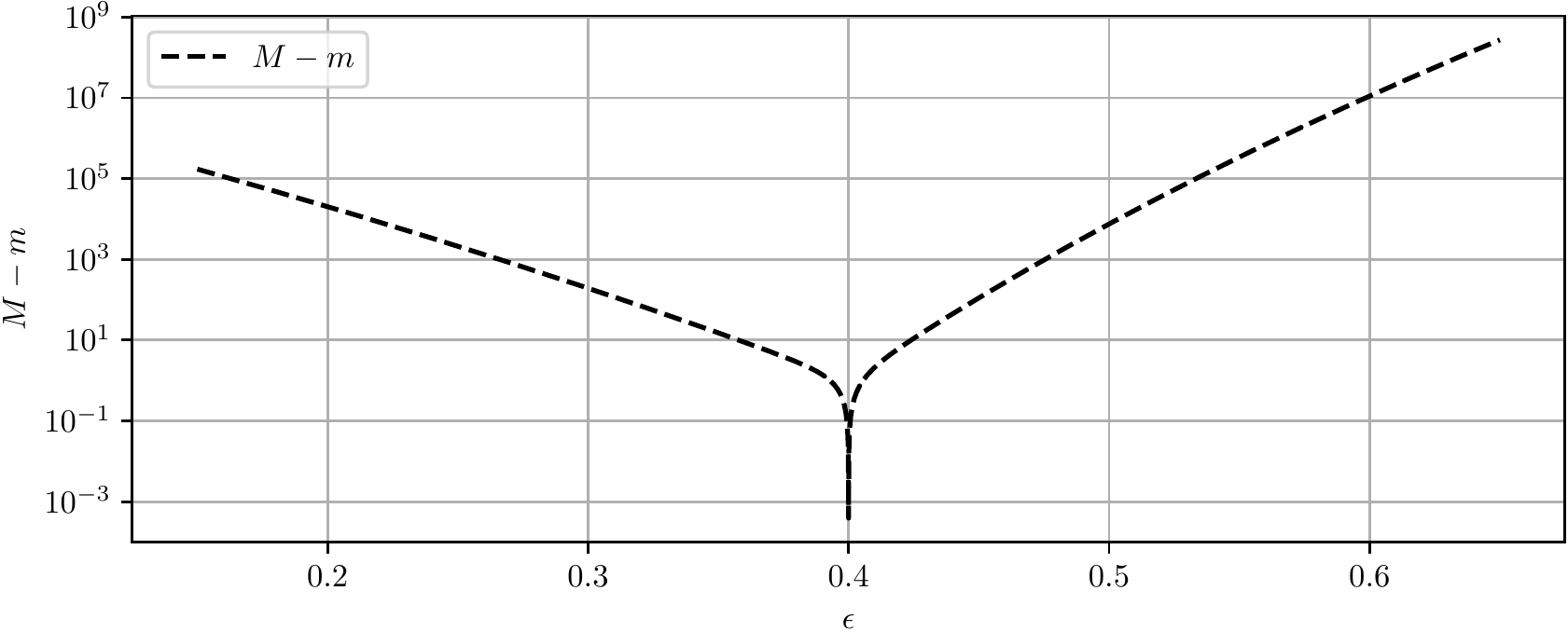}
        \caption{An example of the difference $M-m$ for the case of a convex mixture, with $|{\cal A}|=10, H=40$ and $\epsilon^b=0.4$.}
        \label{fig:scaling_m_M_convex_mixture}
    \end{figure}
    The quantities $M$ and $m$ are usually function of the horizon $H$ (in general one can choose $m=0$). For example, in case ${\cal A}$ is finite, we obtain:
    \begin{itemize}
        \item  If $\pi^b$ is uniform over ${\cal A}$, then an upper bound $M$ is given by $ |{\cal A}|^H $, and $m =  \left(|{\cal A}| \min_{x,a}\pi(a|x)\right)^H$.
        \item In case $\pi^b$ and $\pi$ are convex mixtures of a uniform distribution with another deterministic policy $\hat \pi$, for example $\pi(a|x) = \frac{\epsilon}{|{\cal A}|} + (1-\epsilon)\mathbf{1}_{\{a=\hat\pi(x)\}}$ (sim. $\pi^b$ with $\epsilon^b$), for some $\epsilon,\epsilon^b\geq 0$, then one can choose $M,m$ as 
        \begin{align*}
       M^{1/H}&=\max\left({\epsilon\over \epsilon^b},  \frac{(1-\epsilon)+\epsilon/|{\cal A}|}{(1-\epsilon^b)+\epsilon^b/|{\cal A}|}\right),\\
       m^{1/H} &=\min\left({\epsilon\over \epsilon^b},  \frac{(1-\epsilon)+\epsilon/|{\cal A}|}{(1-\epsilon^b)+\epsilon^b/|{\cal A}|}\right).
        \end{align*}
        See also \cref{fig:scaling_m_M_convex_mixture} for an example of the scaling of $M-m$.
    \end{itemize}

    In general, we see that the dependency on $H$ is mild when $\pi$ and $\pi_b$ that are somehow similar. As a future research direction, we could investigate possible ways to alleviate the impact of $H$ (for example, by looking at the stationary rewards of the MDP, as in \cite{liu2018breaking}).

\medskip
 \subsubsection{Gradient method}\label{paragraph:gradient_method} an alternative approach is to  notice that $w$, as suggested in \cite{taufiq2022conformal}, can be seen as the solution of a MSE minimization problem. Indeed, $w$ solves the following problem:
\begin{equation}
    \min_f \mathbb{E}_{(X,Y,\tau)\sim P_{X,Y,\tau}^{\pi^b}} \left[\left( \frac{  \prod_{t=1}^H \pi(a_t|x_t) }{\prod_{t=1}^H \pi^b(a_t|x_t)} - f(X,Y)\right)^2 \right].
\end{equation}
Therefore, given some function approximator $f_\theta$ parametrized by $\theta$, we can minimize over $\theta$ the following empirical risk:
\[
\frac{1}{m}\sum_{\tau^i \in \datatr}\left( \frac{  \prod_{t=1}^H \pi(a_t^{(i)}|x_t^{(i)}) }{\prod_{t=1}^H \pi^b(a_t^{(i)}|x_t^{(i)})} - f_\theta\left(x_1^{(i)}, \sum_{t=1}^H r_t^{(i)}\right)\right)^2.
\]

As one would expect,  this method  still suffers from a large variance. For large horizons, it becomes quite difficult to learn the ratio of probabilities, especially when the two policies are extremely different. In fact for large $H$, in case the two policies are different, then it is likely that the ratio of action probabilities is $0$ most of the time, with very few values different from $0$ that tend to be extremely large. This makes the training procedure difficult, since most function approximators will just learn to output $0$.

\begin{algorithm}[t]
\caption{Conformal Off-Policy Evaluation in MDPs}\label{alg:OPE_algo}
    \begin{algorithmic}[1]
        \REQUIRE Datasets $\datatr,\datacal$; target coverage $\alpha$; policies $(\pi^b, \pi)$; score function $s$; test input $x^{\text{test}}$.
        %\STATE Compute $w^i = \dfrac{\prod_{t=1}^H \pi(a_t^{(i)}|x_t^{(i)}) }{\prod_{t=1}^H \pi^b(a_t^{(i)}|x_t^{(i)})}$ for $i \in 1, \dots, |\mathcal{D}_{tr}|$
        \STATE Use $\mathcal{D}_{tr}$ to learn the quantiles $q_{\alphalow}(x)$ and $q_{\alphahigh}(x)$, as well as 
        the weight $\hat{w}(x,y)$ using either the empirical estimator or the gradient-based method.
        \STATE Compute $\hat{F}^{x,y}_n$ and the conformalized set $\hat{C}_n(x^{\text{test}})$ using $\hat w(x,y)$ and the scores derived from the dataset $\datacal$ using either (\ref{eq:confidence_set_single_quantile}) or (\ref{eq:confidence_set_double_quantile}) or (\ref{eq:confidence_set_shifted_values2}).\\
        \textbf{Return} $\hat{C}_n(x^{\text{test}})$
    \end{algorithmic}
\end{algorithm}

\subsection{Algorithm}
To conclude this section, we present a generic sketch of our proposed algorithm, see \cref{alg:OPE_algo} for a pseudo-code. Following the split conformal prediction method, the algorithm first leverages the training data $\mathcal{D}_{tr}$ to  estimate the quantiles of the value of $\pi_b$ and the weights $w$. It then uses the calibration data $\mathcal{D}_{cal}$ to compute the non-conformity scores. Using $\hat{w}$ as a plug-in estimate in the re-weighted scores distribution $\hat{F}^{x,y}_n$, the algorithm can finally build the conformal prediction set $\hat{C}_n(x^{\text{test}})$.
\section{Numerical Results}
\ifdefined \technicalreport
\begin{figure*}[t]
  \centering
  \includegraphics[width=.5\linewidth]{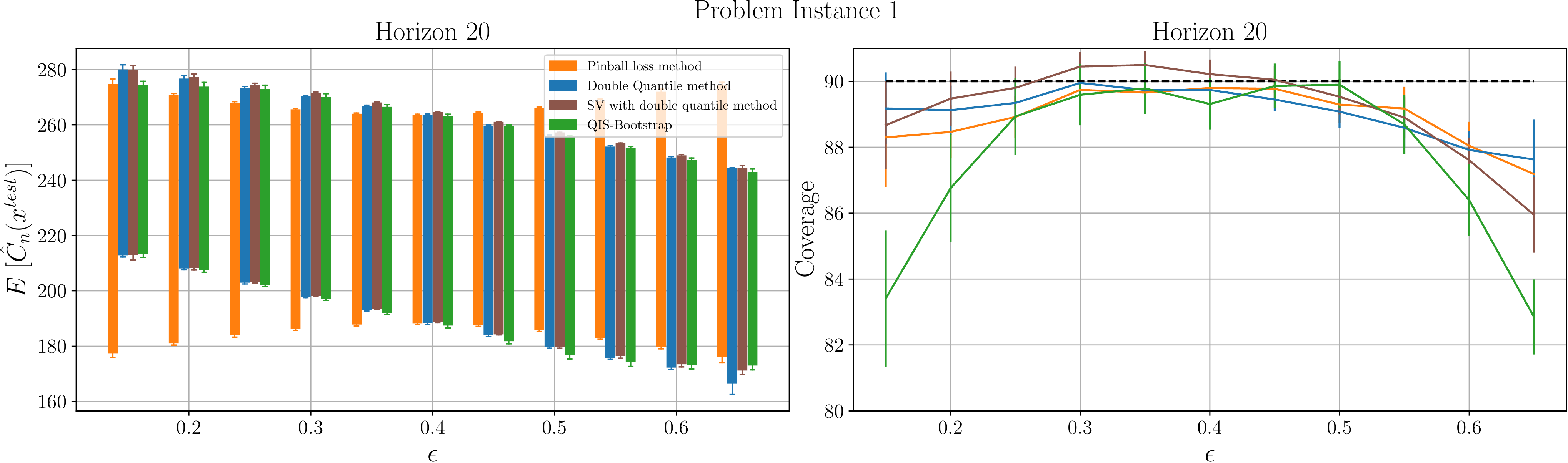}\includegraphics[width=.5\linewidth]{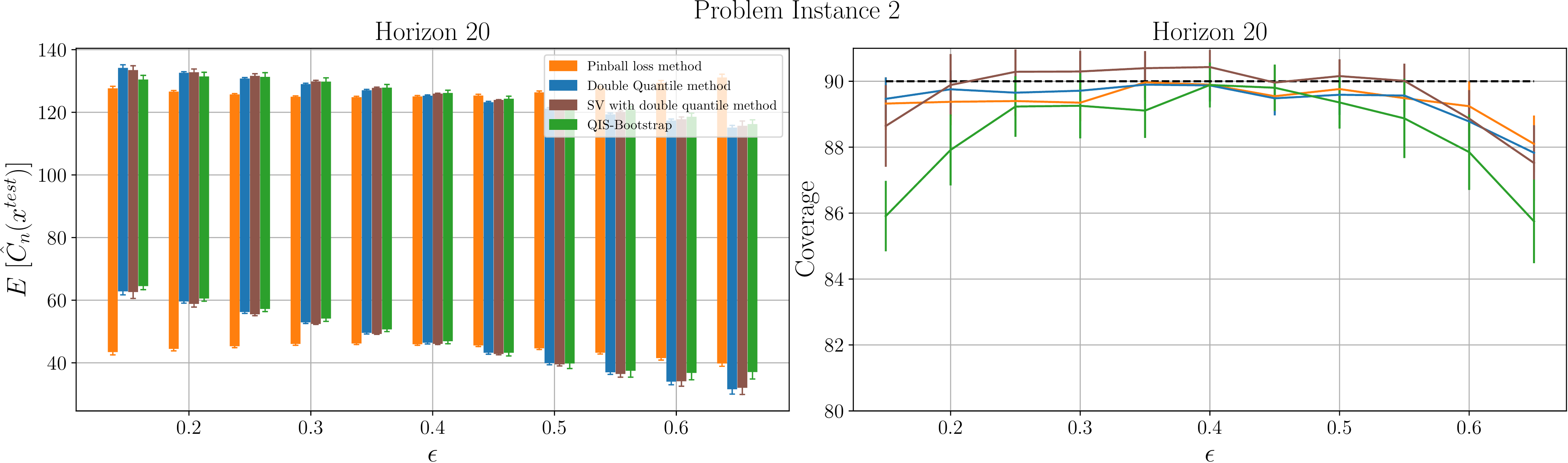}\\[0.3cm]
  \includegraphics[width=.5\linewidth]{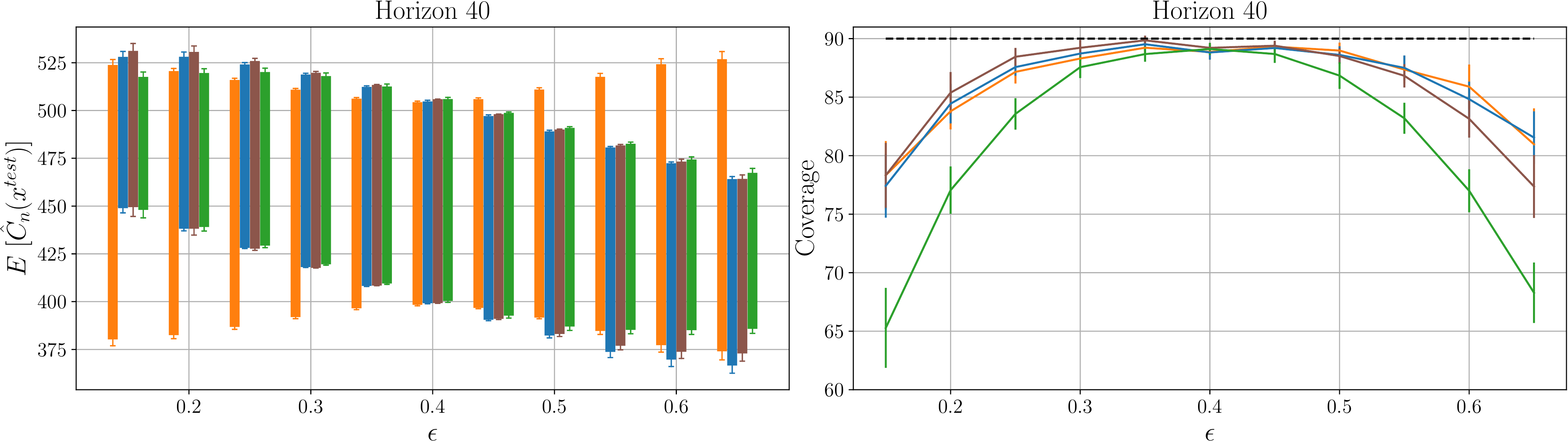}\includegraphics[width=.5\linewidth]{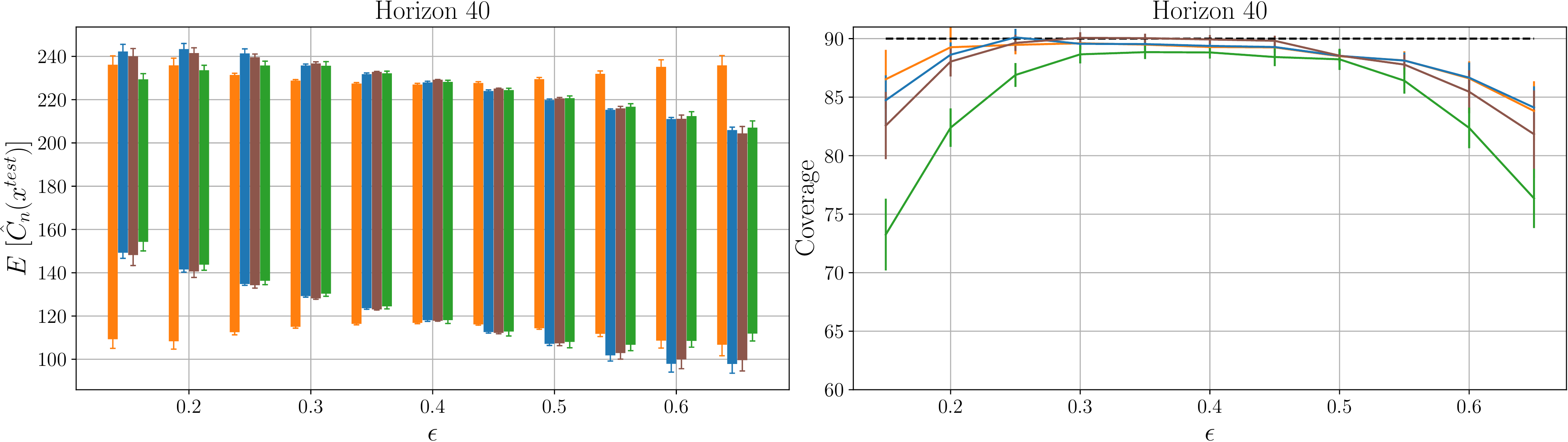}
\caption{Results for the inventory  control problem for $H=20,40$, with target coverage of $90\%$.  The policy $\pi^b$ is $\epsilon^b-$greedy w.r.t. $\pi^\star$ (an optimal discounted policy with discount factor $\gamma=0.99$), with $\epsilon^b = 0.4$.  We evaluated a target policy $\pi$ that is $\epsilon$-greedy w.r.t. $\pi^\star$, with varying $\epsilon$. The four plots on the left are the results corresponding to the first instance of the Inventory Problem, while on the right we present the results for the second instance (both described in section \ref{sec:Environment}). The boxplots show average conformalized intervals for the various methods (whiskers indicate $95\%$ confidence intervals for the minimum and the maximum). The line plots depict the obtained coverage level (bars indicate $95\%$ confidence intervals).}
\label{fig:plots}
\end{figure*}
\else
\begin{figure*}[t]
  \centering
  \includegraphics[width=.95\linewidth]{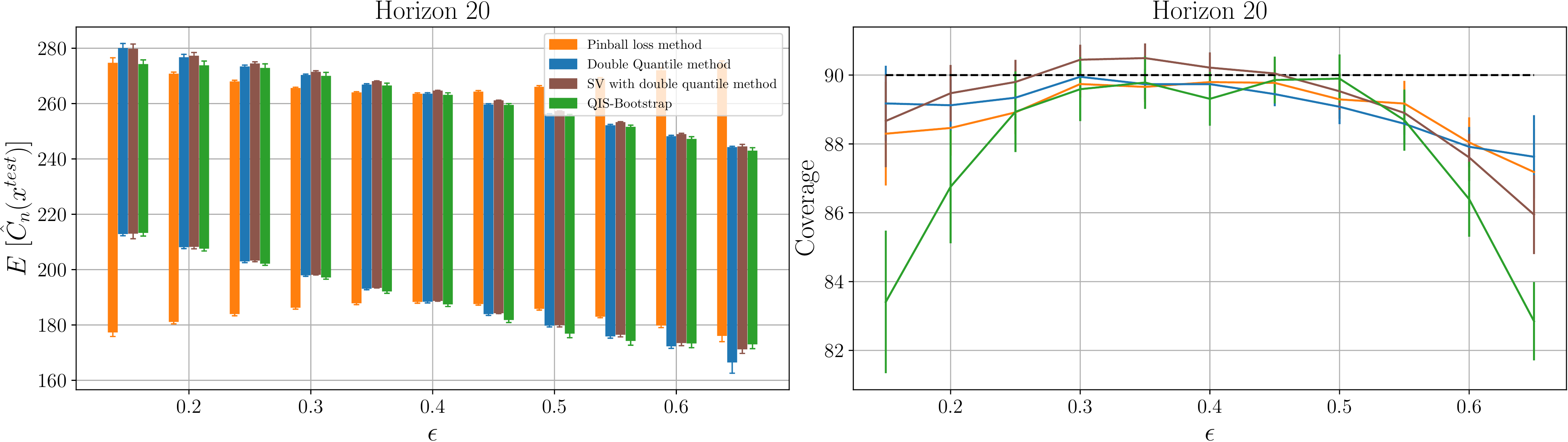}\\[0.5cm]
  \includegraphics[width=.95\linewidth]{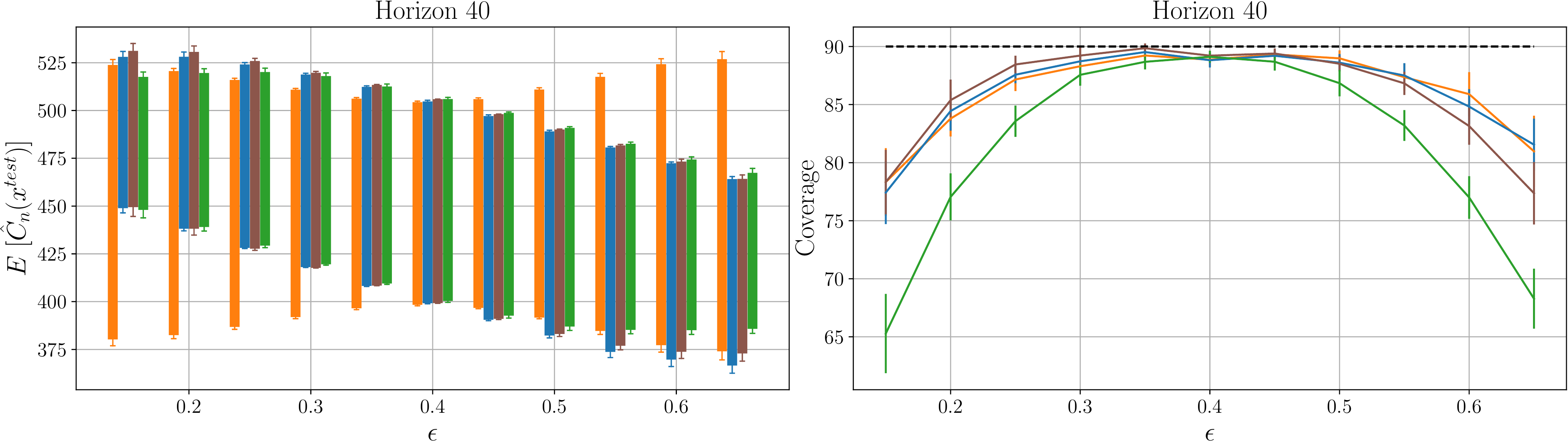}
\caption{Results for the inventory  control problem for $H=20,40$, with target coverage of $90\%$.  The policy $\pi^b$ is $\epsilon^b-$greedy w.r.t. $\pi^\star$ (an optimal discounted policy with discount factor $\gamma=0.99$), with $\epsilon^b = 0.4$.  We evaluated a target policy $\pi$ that is $\epsilon$-greedy w.r.t. $\pi^\star$, with varying $\epsilon$. On the left: we show the boxplots of the average conformalized intervals for the various methods (whiskers indicate $95\%$ confidence intervals for the minimum and the maximum). On the right we depict the coverage (bars indicate $95\%$ confidence intervals).}
\label{fig:plots}
\end{figure*}
\fi

We evaluate our algorithms on the inventory problem \cite{puterman2014markov}, which can be modelled as an MDP with finite state and action spaces. We assume the behavior and target policies $(\pi,\pi^b)$ to be known, and to be $(\epsilon,\epsilon^b)-$greedy with respect to the optimal policy $\pi^\star$. For example, for $\pi$, this means that for all $(x,a)$,
\[
\pi(a|x) = \frac{\epsilon}{|{\cal A}|} + (1-\epsilon)\mathbf{1}_{\{a = \pi^\star(x)\}},
\]
and similarly for $\pi^b$ with $\epsilon^b$. The optimal policy $\pi^\star$ was computed by solving the infinite time-horizon discounted MDP, with discount factor $\gamma = 0.99$.
For each method, we evaluate the prediction interval for the cumulative return of the target policy $\pi$ with different values of $\epsilon$, while the behavior policy  $\pi^b$ has $\epsilon^b = 0.4$. By considering different values of $\epsilon$ for the target policy, we are able to observe how the coverage and interval length vary with respect to the distance between the target and the behavior policies.

\subsection{Environment}
\label{sec:Environment}
The inventory control problem is modelled as follows: an agent manages an inventory of size $N$ while facing a stochastic demand for what is stored in it. At each round, the agent must choose how many items to buy to meet the upcoming order for the next day. The action set is the same for every state, i.e. $\mathcal{A} = \left[0,N\right]$. We define the cost of buying $a$ items as $k\mathbf{1}_{\{a>0\}} + c (\min(N,x_t + a)- x_t)$, where $k>0$ is the fixed cost for a single order and $c>0$ is the cost of a single unit bought. At each round, the agent earns a quantity $pl$, where $p$ is the price of a single item and $l$ is the number of items sold. Finally, the agent has to pay a cost $zn$ for storing $n > 0$ items, with $z > 0$ and $p > z$. The order $o_t$ is sampled from a Poisson distribution with rate $\lambda$. The next state is computed according to $x_{t+1} = \max(0, \min(N, x_t + a_t) - o_{t+1})$, while the reward is the sum of the costs and earnings listed above, i.e., $r(x_t, a_t, x_{t+1}) = -k\mathbf{1}_{\{a_t>0\}} - zx_t -c (\min(N,x_t + a_t)- x_t) + p \max(0, \min(N, x_t + a_t)-x_{t+1})$. Note that here, the rewards are deterministic but depend on the next state -- we can easily verify that all our results naturally extend to this setting.
\ifdefined \technicalreport
We considered two instances of the inventory environment when evaluating our algorithm. In the first one we chose the following parameters: $N = 10$, $k = 1$, $c = 2$, $z = 2$, $p = 4$, $\lambda = 10$. For the second one, we modified the parameters such that: $k = 3$, $\lambda = 6$. So in the second instance, the agent was penalized more for making a single order (i.e., higher $k$) and the demand rate $\lambda$ was decreased.
\else
When testing our algorithm we chose the following parameters: $N = 10$, $k = 1$, $c = 2$, $z = 2$, $p = 4$, $\lambda = 10$. 
\fi

\subsection{Algorithm details}

We consider three different implementations of our algorithm: the first using the classical pinball score function \cite{romano2019conformalized, taufiq2022conformal}, the second using the double quantile method and finally the shifted values method with double quantile.

\subsubsection{Pinball score function}\label{subsec:original_method}
this method adapts the algorithm presented in \cite{taufiq2022conformal} to our setting (which is described in \cref{subsec:score_function}). We use the training dataset $\mathcal{D}_{tr}$ to also learn two quantile networks $\hat{q}_{\alphalow}$ and $\hat{q}_{\alphahigh}$, with $\alphalow = \alpha/2,\alphahigh=1-\alpha/2$ (where $\alpha$ is the coverage parameter). The two functions are estimated using quantile regression and are modelled using two neural networks with two hidden layers of $64$ nodes and ReLU activation functions. 
For this approach, the score function used is $s(x,y) = \max\left( y - \hat{q}_{\alphahigh}, \hat{q}_{\alphalow} - y\right).$ Once we have computed the empirical CDF of the scores $\hat{F}_n^{x,y}$, the confidence set is obtained using (\ref{eq:confidence_set_single_quantile}).

\subsubsection{Double Quantile (DQ) method}\label{subsec:double_quantile_method}
Here we apply the method in \ref{paragraph:double_quantile}. In this method, we introduce two score functions
\begin{align*}
    s_0(x,y)&=\hat q_{\alphalow}(x) -y\\ 
    s_1(x,y)&= y-\hat q_{\alphahigh}(x),
\end{align*}
where $\hat{q}_{\alphalow}$ and $\hat{q}_{\alphahigh}$ are the same networks as in the previous method. Lastly, the confidence set is computed using (\ref{eq:confidence_set_double_quantile}).

\subsubsection{Shifted Values (SV) with double quantile method}\label{subsec:shifted_values_method}
Here we consider a score function that allows us to shift the values of the behavior policy $s(x,y)=y$, as explained in \ref{paragraph:shifted_values}, and compute the confidence set according to (\ref{eq:confidence_set_shifted_values2}).

\subsection{Baseline: Quantile Estimation through Importance Sampling with Bootstrap (QIS-Bootstrap)}\label{subsec:baseline_quantile_bootstrap}
We compare the conformal prediction method developed in this work to quantile estimation through importance sampling  \cite{glynn1996importance} with bootstrap. Importance sampling (IS) has been widely used as a variance reduction technique in statistical methods, but in our case it can be used to perform off-policy evaluation as in \cite{precup2000eligibility,thomas2015highimprovement}. However, compared to \cite{precup2000eligibility,thomas2015highimprovement} that try to estimate the mean value of the target policy $\pi$, we use the IS technique to estimate the $(\alphalow,\alphahigh)$-quantiles of the value of $\pi$.
The key insight is that  $q_\alpha^\pi(x)$, the $\alpha$-quantile of $\pi$ in $x$, can be estimated using the calibration data $\datacal$ and the likelihood ratio $w(x,y)$ through the following expression
\begin{equation*}
  q_\alpha^\pi(x) = \quantile{\alpha}\left(\sum_{y\in {\cal I}(x)}  \frac{w\left(x,y\right)}{\sum_{y'\in {\cal I}(x)}  w\left(x,y'\right)}\delta_y \right),
\end{equation*}
where ${\cal I}(x)=\{y \in \datacal: x_1=x\}$, \emph{i.e.}, we only consider the cumulative rewards of the trajectories in $\datacal$ that start in $x$.

The inner term can be seen as an empirical estimator of $F_x^\pi(y)=\mathbb{E}_{Y\sim P^\pi(Y|X=x)}[\mathbf{1}_{\{Y\leq y\}}]=\mathbb{E}_{Y\sim P^{\pi^b}(Y|X=x)}[w(x,Y)\mathbf{1}_{\{Y\leq y\}}]$, the CDF of the values of $\pi$ in $x$ (note that the normalization factor does not affect the outcome, see also \cite{glynn1996importance}). Since $w(x,y)$ is unknown, we replace it by  $\hat w(x,y)$.

Next, rather than using the estimate $q_\alpha^\pi$ directly, to obtain a better estimate we  use bootstrapping \cite{tibshirani1993introduction} to estimate a confidence interval around the $\alpha$-quantile, obtaining a high-confidence interval $(q_{\alpha-}^\pi, q_{\alpha+}^\pi)$ and then taking the median point $\bar q_\alpha(x)\coloneqq (q_{\alpha-}^\pi + q_{\alpha+}^\pi)/2$. Finally, the confidence set for the value of $\pi$ is simply given by
\begin{equation}
    \hat C_n(x) = [\bar q_{\alphalow}(x), \bar q_{\alphahigh}(x)].
\end{equation}
It is important to remember that there is no coverage guarantees for this set $\hat C_n(x)$.

\subsection{Results and discussion}
In Figure \ref{fig:plots}, we show the results of our methods in the Inventory Problem
\ifdefined \technicalreport
 for horizons 20 and 40 (for both the problem instances defined in \ref{sec:Environment}),
 \else
for horizons 20 and 40,
\fi 
where results are averaged over 30 runs. Recall that the policy $\pi^b$ is $\epsilon^b-$greedy w.r.t. $\pi^\star$, with $\epsilon^b = 0.4$, while $\pi$ is $\epsilon$-greedy w.r.t $\pi^\star$, with $\epsilon$ varying in $[0.15, 0.65]$.
The target level of coverage was chosen  as $1-\alpha=90\%$ (depicted as the dashed black line in the plots of the second column). We evaluated our algorithms using the empirical estimate of $\hat{w}$ (see \ref{paragraph:empirical_estimator}) against the QIS-Bootstrap baseline method  in \cref{subsec:baseline_quantile_bootstrap}.

\subsubsection{Conformalized intervals} 
\ifdefined \technicalreport
the boxplots illustrate the conformalized interval obtained for each method.
\else
the plots in the left column illustrate the conformalized interval obtained for each method as a boxplot.
\fi
For each run, method, and value of $\epsilon$, we evaluated the confidence interval across $2000$ tests-points $x^{\text{test}}$ sampled from $p(x)$, and averaged the corresponding minimum and maximum values of the confidence set $\hat C_n(x^{\text{test}})$. The whiskers indicate $95\%$ confidence interval for the minimum and the maximum. As mentioned in \cref{subsec:score_function}, we observe that the pinball method yields an interval that enlarges/shrinks symmetrically around a fixed point. As a consequence, the interval becomes larger to maintain the desired coverage when the target policy $\pi$ becomes really different than $\pi^b$ ({i.e.}, $\epsilon$ is different than $\epsilon^b=0.4$). Instead, with the proposed double quantile method, the interval is shifted depending on how far the target policy $\pi$ is w.r.t. $\pi^b$, leading to smaller intervals even when the policies are far from each other. The intervals estimated by the QIS-Bootstrap method match the ones of our new score functions when $\pi$ is close to $\pi^b$. However, when the policies are far from each other, the estimated interval is too conservative (i.e., too small and off-centred), which reflects in the coverage level of the algorithm, quickly degrading as $\pi$ moves away from $\pi^b$, for both horizons.

\subsubsection{Coverage} 
\ifdefined \technicalreport
the line plots illustrate the achieved coverage, averaged over 30 runs (bars indicate $95\%$ confidence interval).
\else
the plots in the right column illustrate the achieved coverage, averaged over 30 runs (bars indicate $95\%$ confidence interval).
\fi
All the proposed conformalized methods achieved better levels of coverage than QIS-Bootstrap, as one would expect. For horizon $H=20$, the pinball method can maintain the desired level of coverage for all the epsilons at the expense of the interval length, while the new methods achieve a better level of coverage with a smaller interval size. For a larger horizon ($H = 40$), we can see that the coverage of the QIS-Bootstrap method degrades very rapidly, maintaining the desired level only for  $\pi\approx\pi^b$.

\subsubsection{Discussion and future work}\label{sec:future} Some of the methods discussed to estimate the likelihood ratio $w(x,y)$ were not used in our numerical experiments. This is mostly due to computational challenges: as we previously mentioned, the computational complexity of the  Monte-Carlo method vastly exceeds the complexity of the other methods (empirical estimator and gradient method), while the gradient method has several difficulties in learning the likelihood ratios for values of $(\epsilon,\epsilon^b)$ that greatly differ. We plan to investigate how to efficiently learn the likelihood ratios using neural networks. Finally, we note that one may try conformalize the QIS-Bootstrap method in \cref{subsec:baseline_quantile_bootstrap} to have a more fair comparison.

\section{Conclusion}
In this work, we considered the \emph{offline off-policy evaluation} problem in finite time-horizon Markov Decision Processes. Using Conformal Prediction (CP) techniques, we developed methods to construct conformalized intervals that include the true reward of the target policy with a prescribed level of certainty. Some of the challenges addressed in this paper include dealing with time-dependent data, as well as addressing the distribution shift between the behavior policy and the target policy. Furthermore, we proposed improved CP methods that allow to obtain intervals  with significantly reduced length when compared to existing CP methods, while retaining the same certainty guarantees. We conclude with numerical results on the inventory control problem that demonstrated the efficiency of our methods. Several interesting research directions have been mentioned in the text, of which, the most significant, consists in improving the estimation of the likelihood ratio characterizing the distribution shift.


%\newpage

%\input{B.acknowledgment.tex}

\bibliographystyle{plain} % We choose the "plain" reference style
\bibliography{refs} % Entries are in the refs.bib file

\begin{thebibliography}{10}

\bibitem{bottou2013counterfactual}
L{\'e}on Bottou, Jonas Peters, Joaquin Qui{\~n}onero-Candela, Denis~X Charles,
  D~Max Chickering, Elon Portugaly, Dipankar Ray, Patrice Simard, and
  Ed~Snelson.
\newblock Counterfactual reasoning and learning systems: The example of
  computational advertising.
\newblock {\em Journal of Machine Learning Research}, 14(11), 2013.

\bibitem{dietterich2022conformal}
Thomas~G Dietterich and Jesse Hostetler.
\newblock Conformal prediction intervals for markov decision process
  trajectories.
\newblock {\em arXiv preprint arXiv:2206.04860}, 2022.

\bibitem{duan2020minimax}
Yaqi Duan, Zeyu Jia, and Mengdi Wang.
\newblock Minimax-optimal off-policy evaluation with linear function
  approximation.
\newblock In {\em International Conference on Machine Learning}, pages
  2701--2709. PMLR, 2020.

\bibitem{glynn1996importance}
Peter~W Glynn et~al.
\newblock Importance sampling for monte carlo estimation of quantiles.
\newblock In {\em Mathematical Methods in Stochastic Simulation and
  Experimental Design: Proceedings of the 2nd St. Petersburg Workshop on
  Simulation}, pages 180--185. Citeseer, 1996.

\bibitem{hanna2017bootstrapping}
Josiah Hanna, Peter Stone, and Scott Niekum.
\newblock Bootstrapping with models: Confidence intervals for off-policy
  evaluation.
\newblock In {\em Proceedings of the AAAI Conference on Artificial
  Intelligence}, volume~31, 2017.

\bibitem{jiang2020minimax}
Nan Jiang and Jiawei Huang.
\newblock Minimax value interval for off-policy evaluation and policy
  optimization.
\newblock {\em Advances in Neural Information Processing Systems},
  33:2747--2758, 2020.

\bibitem{kallus2020double}
Nathan Kallus and Masatoshi Uehara.
\newblock Double reinforcement learning for efficient off-policy evaluation in
  markov decision processes.
\newblock {\em The Journal of Machine Learning Research}, 21(1):6742--6804,
  2020.

\bibitem{kallus2022efficiently}
Nathan Kallus and Masatoshi Uehara.
\newblock Efficiently breaking the curse of horizon in off-policy evaluation
  with double reinforcement learning.
\newblock {\em Operations Research}, 2022.

\bibitem{kuzborskij2021confident}
Ilja Kuzborskij, Claire Vernade, Andras Gyorgy, and Csaba Szepesv{\'a}ri.
\newblock Confident off-policy evaluation and selection through self-normalized
  importance weighting.
\newblock In {\em International Conference on Artificial Intelligence and
  Statistics}, pages 640--648. PMLR, 2021.

\bibitem{le2019batch}
Hoang Le, Cameron Voloshin, and Yisong Yue.
\newblock Batch policy learning under constraints.
\newblock In {\em International Conference on Machine Learning}, pages
  3703--3712. PMLR, 2019.

\bibitem{lei2014distribution}
Jing Lei and Larry Wasserman.
\newblock Distribution-free prediction bands for non-parametric regression.
\newblock {\em Journal of the Royal Statistical Society: Series B (Statistical
  Methodology)}, 76(1):71--96, 2014.

\bibitem{lei2021conformal}
Lihua Lei and Emmanuel~J Cand{\`e}s.
\newblock Conformal inference of counterfactuals and individual treatment
  effects.
\newblock {\em Journal of the Royal Statistical Society Series B: Statistical
  Methodology}, 83(5):911--938, 2021.

\bibitem{lindemann2022safe}
Lars Lindemann, Matthew Cleaveland, Gihyun Shim, and George~J Pappas.
\newblock Safe planning in dynamic environments using conformal prediction.
\newblock {\em arXiv preprint arXiv:2210.10254}, 2022.

\bibitem{lindh2017predicting}
Martin Lindh, Anders Karl{\'e}n, and Ulf Norinder.
\newblock Predicting the rate of skin penetration using an aggregated conformal
  prediction framework.
\newblock {\em Molecular Pharmaceutics}, 14(5):1571--1576, 2017.

\bibitem{liu2018breaking}
Qiang Liu, Lihong Li, Ziyang Tang, and Dengyong Zhou.
\newblock Breaking the curse of horizon: Infinite-horizon off-policy
  estimation.
\newblock {\em Advances in Neural Information Processing Systems}, 31, 2018.

\bibitem{lu2022three}
Charles Lu, Ken Chang, Praveer Singh, and Jayashree Kalpathy-Cramer.
\newblock Three applications of conformal prediction for rating breast density
  in mammography.
\newblock {\em arXiv preprint arXiv:2206.12008}, 2022.

\bibitem{manokhinvalery20226467205}
Valery Manokhin.
\newblock Awesome conformal prediction, April 2022.
\newblock {"If you use Awesome Conformal Prediction, please cite it as
  below."}.

\bibitem{papadopoulos2002inductive}
Harris Papadopoulos, Kostas Proedrou, Volodya Vovk, and Alex Gammerman.
\newblock Inductive confidence machines for regression.
\newblock In {\em European Conference on Machine Learning}, pages 345--356.
  Springer, 2002.

\bibitem{precup2000eligibility}
Doina Precup.
\newblock Eligibility traces for off-policy policy evaluation.
\newblock {\em Computer Science Department Faculty Publication Series},
  page~80, 2000.

\bibitem{puterman2014markov}
Martin~L Puterman.
\newblock {\em Markov decision processes: discrete stochastic dynamic
  programming}.
\newblock John Wiley \& Sons, 2014.

\bibitem{romano2019conformalized}
Yaniv Romano, Evan Patterson, and Emmanuel Candes.
\newblock Conformalized quantile regression.
\newblock {\em Advances in neural information processing systems}, 32, 2019.

\bibitem{shi2021deeply}
Chengchun Shi, Runzhe Wan, Victor Chernozhukov, and Rui Song.
\newblock Deeply-debiased off-policy interval estimation.
\newblock In {\em International Conference on Machine Learning}, pages
  9580--9591. PMLR, 2021.

\bibitem{shi2022statistical}
Chengchun Shi, Sheng Zhang, Wenbin Lu, and Rui Song.
\newblock Statistical inference of the value function for reinforcement
  learning in infinite-horizon settings.
\newblock {\em Journal of the Royal Statistical Society Series B: Statistical
  Methodology}, 84(3):765--793, 2022.

\bibitem{taufiq2022conformal}
Muhammad~Faaiz Taufiq, Jean-Francois Ton, Rob Cornish, Yee~Whye Teh, and Arnaud
  Doucet.
\newblock Conformal off-policy prediction in contextual bandits.
\newblock {\em arXiv preprint arXiv:2206.04405}, 2022.

\bibitem{thomas2015highevaluation}
Philip Thomas, Georgios Theocharous, and Mohammad Ghavamzadeh.
\newblock High-confidence off-policy evaluation.
\newblock In {\em Proceedings of the AAAI Conference on Artificial
  Intelligence}, volume~29, 2015.

\bibitem{thomas2015highimprovement}
Philip Thomas, Georgios Theocharous, and Mohammad Ghavamzadeh.
\newblock High confidence policy improvement.
\newblock In {\em International Conference on Machine Learning}, pages
  2380--2388. PMLR, 2015.

\bibitem{tibshirani1993introduction}
Robert~J Tibshirani and Bradley Efron.
\newblock An introduction to the bootstrap.
\newblock {\em Monographs on statistics and applied probability}, 57(1), 1993.

\bibitem{tibshirani2019conformal}
Ryan~J Tibshirani, Rina Foygel~Barber, Emmanuel Candes, and Aaditya Ramdas.
\newblock Conformal prediction under covariate shift.
\newblock {\em Advances in neural information processing systems}, 32, 2019.

\bibitem{uehara2022review}
Masatoshi Uehara, Chengchun Shi, and Nathan Kallus.
\newblock A review of off-policy evaluation in reinforcement learning, 2022.

\bibitem{vovk2005algorithmic}
Vladimir Vovk, Alexander Gammerman, and Glenn Shafer.
\newblock {\em Algorithmic learning in a random world}.
\newblock Springer Science \& Business Media, 2005.

\bibitem{wisniewski2020application}
Wojciech Wisniewski, David Lindsay, and Sian Lindsay.
\newblock Application of conformal prediction interval estimations to market
  makers’ net positions.
\newblock In {\em Conformal and Probabilistic Prediction and Applications},
  pages 285--301. PMLR, 2020.

\bibitem{xi2022conformal}
Zepu Xi, Xuebin Zhuang, and Hongbo Chen.
\newblock Conformal prediction for hypersonic flight vehicle classification.
\newblock In {\em Conformal and Probabilistic Prediction with Applications},
  pages 118--206. PMLR, 2022.

\bibitem{zhan2020electronic}
Xianghao Zhan, Zhan Wang, Meng Yang, Zhiyuan Luo, You Wang, and Guang Li.
\newblock An electronic nose-based assistive diagnostic prototype for lung
  cancer detection with conformal prediction.
\newblock {\em Measurement}, 158:107588, 2020.

\end{thebibliography}

\end{document}